\pdfoutput=1

\documentclass[11pt]{article}
\usepackage{graphicx}
\usepackage[]{EMNLP2022}
\usepackage{todonotes}
\usepackage{times}
\usepackage{latexsym}
\usepackage{adjustbox}
\usepackage{xspace}
\usepackage[T1]{fontenc}
\usepackage{graphicx}
\usepackage{subcaption}
\usepackage[utf8]{inputenc}

\usepackage{microtype}

\usepackage{inconsolata}

\usepackage{multirow}
\makeatletter
\newcommand\HUGE{\@setfontsize\Huge{50}{60}}
\makeatother

\title{Position Matters! Empirical Study of Order Effect  \\
in Knowledge-grounded Dialogue}

\author{
Hsuan Su$^{\star\diamond}$\thanks{\xspace~Work done when interning at Intel Labs.}  \, Shachi H Kumar$^{\diamond}$ \, Sahisnu Mazumder$^{\diamond}$ \, Wenda Chen$^{\diamond}$\\ 
\textbf{ Ramesh Manuvinakurike$^{\diamond}$} \, \textbf{Eda Okur$^{\diamond}$} \, \textbf{Saurav Sahay$^{\diamond}$} \, \textbf{Lama Nachman$^{\diamond}$} \, \textbf{Shang-Tse Chen$^\star$} \, \textbf{Hung-yi Lee$^\star$} \\
        {National Taiwan University$^\star$} \, {Intel Labs$^{\diamond}$} \\
        \small hsuansu.96@gmail.com
        }
\begin{document}
\maketitle
\begin{abstract}
With the power of large pretrained language models, various research works have integrated knowledge into dialogue systems. The traditional techniques treat knowledge as part of the input sequence for the dialogue system, prepending a set of knowledge statements in front of dialogue history.
However, such a mechanism forces knowledge sets to be concatenated in an ordered manner, making models implicitly pay imbalanced attention to the sets during training.
In this paper, we first investigate how the order of the knowledge set can influence autoregressive dialogue systems' responses. 
We conduct experiments on two commonly used dialogue datasets with two types of transformer-based models and find that models view the input knowledge unequally. 
To this end, we propose a simple and novel technique to alleviate the order effect by modifying the position embeddings of knowledge input in these models. With the proposed position embedding method, the experimental results show that each knowledge statement is uniformly considered to generate responses.
\end{abstract}

\section{Introduction}

Transformer-based \cite{NIPS2017_3f5ee243} pretrained language models are widely used to build dialogue systems \cite{zhang2019dialogpt, https://doi.org/10.48550/arxiv.2107.07567, https://doi.org/10.48550/arxiv.2107.07566, https://doi.org/10.48550/arxiv.2004.13637, https://doi.org/10.48550/arxiv.2201.08239, https://doi.org/10.48550/arxiv.2112.11446, chen2021teaching, ham-etal-2020-end, hosseini2020simple, Bao2021AUP}. In addition to general-purpose dialogue systems, many specialized dialogue systems have been proposed. Representative examples include personalized dialogue systems \cite{DBLP:journals/corr/abs-1901-08149, zhang-etal-2018-personalizing, wu-etal-2021-transferable, Cao2022AMD, Song2020GenerateDA}, knowledge-grounded dialogue systems \cite{Dinan2019WizardOW, Kim2021AMO, Tao2021APS, Cai2020ABT, Liu2021ATL}, and prompting dialogue systems \cite{su2022fewshot}.

\begin{figure}
    \centering
    \includegraphics[width = \linewidth]{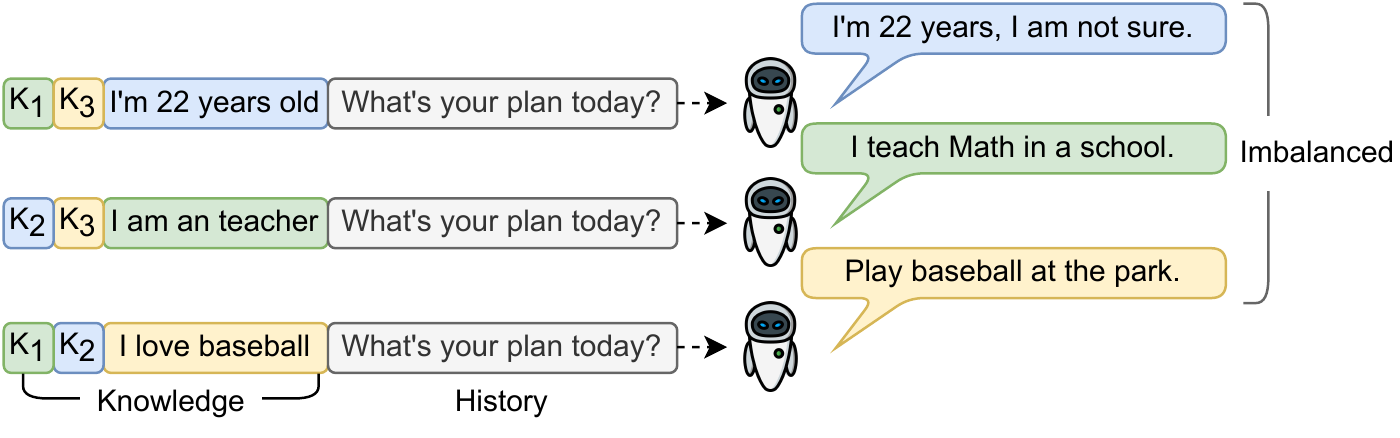}
    \caption{The order effect illustration. Models' responses are influenced by the order of the input knowledge set.}
    \label{fig:example}
\end{figure}

To build specialized dialogue systems, integrating additional information into the input sequence is necessary. \citet{DBLP:journals/corr/abs-1901-08149} prepend persona sentences to personalize the history; while \citet{su2022fewshot, dinan-etal-2020-queens, keskar2019ctrl, xu-etal-2020-megatron} prepending task-specific signals to prompt and control the model.

\begin{figure*}[htp]
    \centering
    \includegraphics[width = \linewidth]{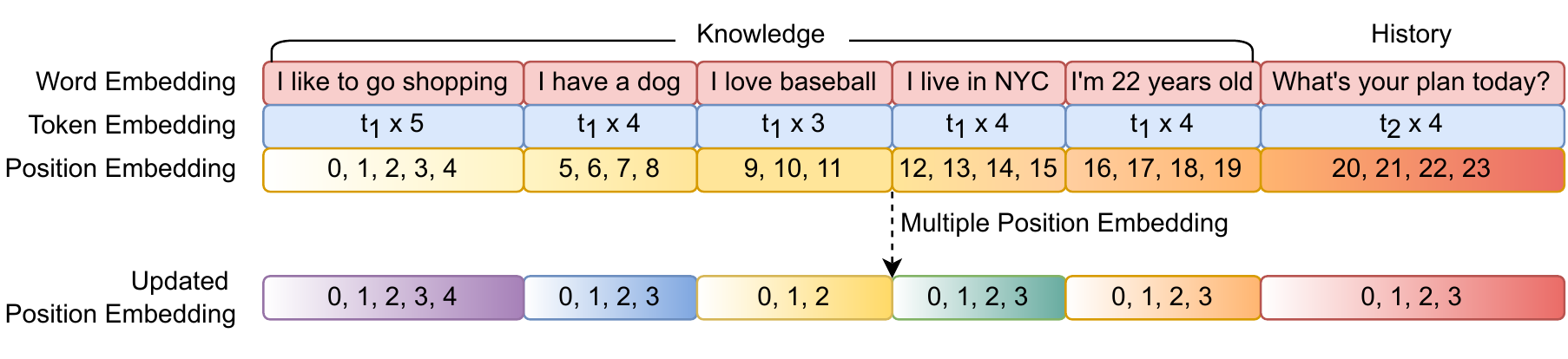}
    \caption{Input format for GPT-series models. The position ids do not treat knowledge equally but as a sequence. 
    The updated position embeddings show our proposed method, where each knowledge statement is encoded with its own position embeddings, hence, models can treat each input sentence equally during training. The same color of blocks indicates using the same layer to generate embeddings.}
    \label{fig:framework}
\end{figure*}
These methods prepend additional information in front of the history as a sequence for models' input. Furthermore, the approach generates an unnecessary order among equal knowledge sets since the knowledge is connected in the sequence. Thus models might be influenced by the order and generate imbalanced responses.

 Previous works focus on how perturbations in dialog history affect models' responses \cite{sankar-etal-2019-neural, oconnor-andreas-2021-context, sinha-etal-2021-masked, Lampinen2022CanLM, Webson2021DoPM, Xu2020ATO, Khandelwal2018SharpNF}. They conduct many experiments and measure the effect of perturbations from the aspect of response quality and information theory to show that these language models are robust and not sensitive to the perturbations in input history.
However, dialog history and knowledge are inherently different aspects of a conversation. Dialog history has a temporal property, i.e., the topic and specificity of conversation change as the dialog progresses, whereas knowledge facts are information referenced to generate a response. Although the perturbation in history does not influence the results generated by the model \cite{sankar-etal-2019-neural, oconnor-andreas-2021-context}, 
in our early observation, we found that prepending knowledge influences models' responses.
For example, Figure \ref{fig:example} demonstrates an example where the model exhibits imbalanced attention to input knowledge, and the order of knowledge influences the generated responses. This might cause the model to generate inappropriate responses since it attends to knowledge that might not be relevant to a dialog context.
The contributions of this work are as follows:
\begin{itemize}
    \item We conduct experiments across two typical methods and two models on multiple datasets to show that the order of knowledge sentences does affect generated responses. 
    \item We propose a simple approach to alleviate this sentence-level order effect by manipulating the position embedding layers.
\end{itemize}


\section{Knowledge-grounded Dialogue Methods}
In this work, we study the order effect in TransferTransfo \cite{DBLP:journals/corr/abs-1901-08149}, which is a state-of-the-art knowledge-grounded method.
We train TransferTransfo on two datasets and measure the sentence-level order effect on the test datasets.

\subsection{TransferTransfo}

The TransferTransfo architecture is built on top of GPT-series models, which simply concatenates the knowledge sets and context in a single sequence, putting the reply at the end. To help models distinguish speakers and position of input tokens, it builds three parallel input sequences for word, position, and segments, and fuses them into a single sequence.
For the loss function, in addition to a language modeling loss, a next sentence prediction loss is added. 
The total loss is the weighted sum of the 1) language modeling loss, which is computed as the cross-entropy loss between the predicted logits and the ground truth response and 2) the next-sentence prediction loss, which is a classification loss to distinguish the ground truth response from distractors that are randomly sampled from the dataset.

In the original TransferTransfo implementation, the authors have already pointed out that the order of the knowledge set influences the model's performance. To this end, they augment training data by permuting the knowledge sets several times. 

    \begin{figure*}[t]
        \centering
        \begin{subfigure}[b]{0.245\linewidth}
            \centering
            \includegraphics[width=\linewidth]{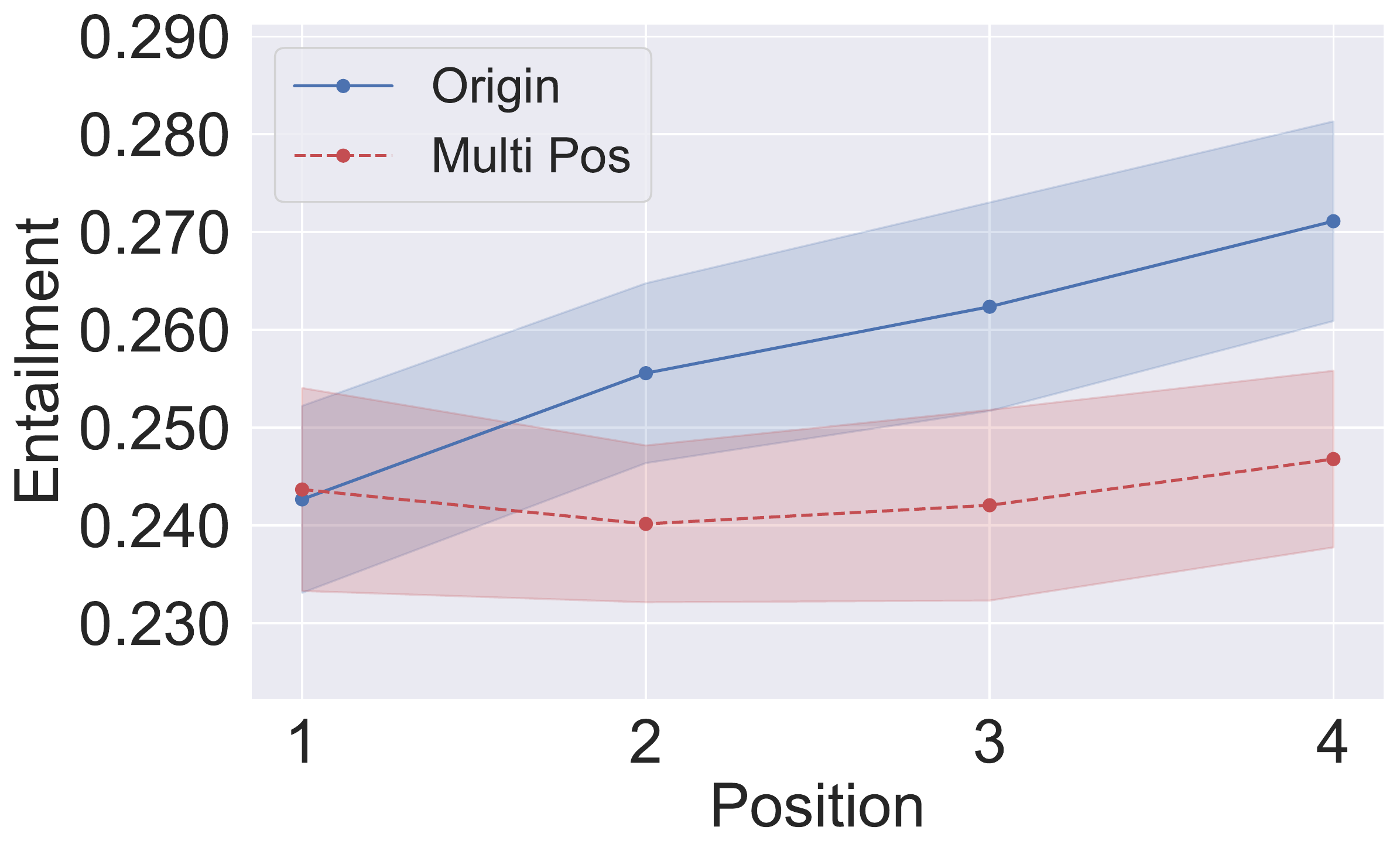}
            \caption[]%
            {{\tiny GPT on Persona-Chat}}    
            \label{fig:gpt_4_p}
        \end{subfigure}
        \hfill
        \begin{subfigure}[b]{0.245\linewidth}  
            \centering 
            \includegraphics[width=\linewidth]{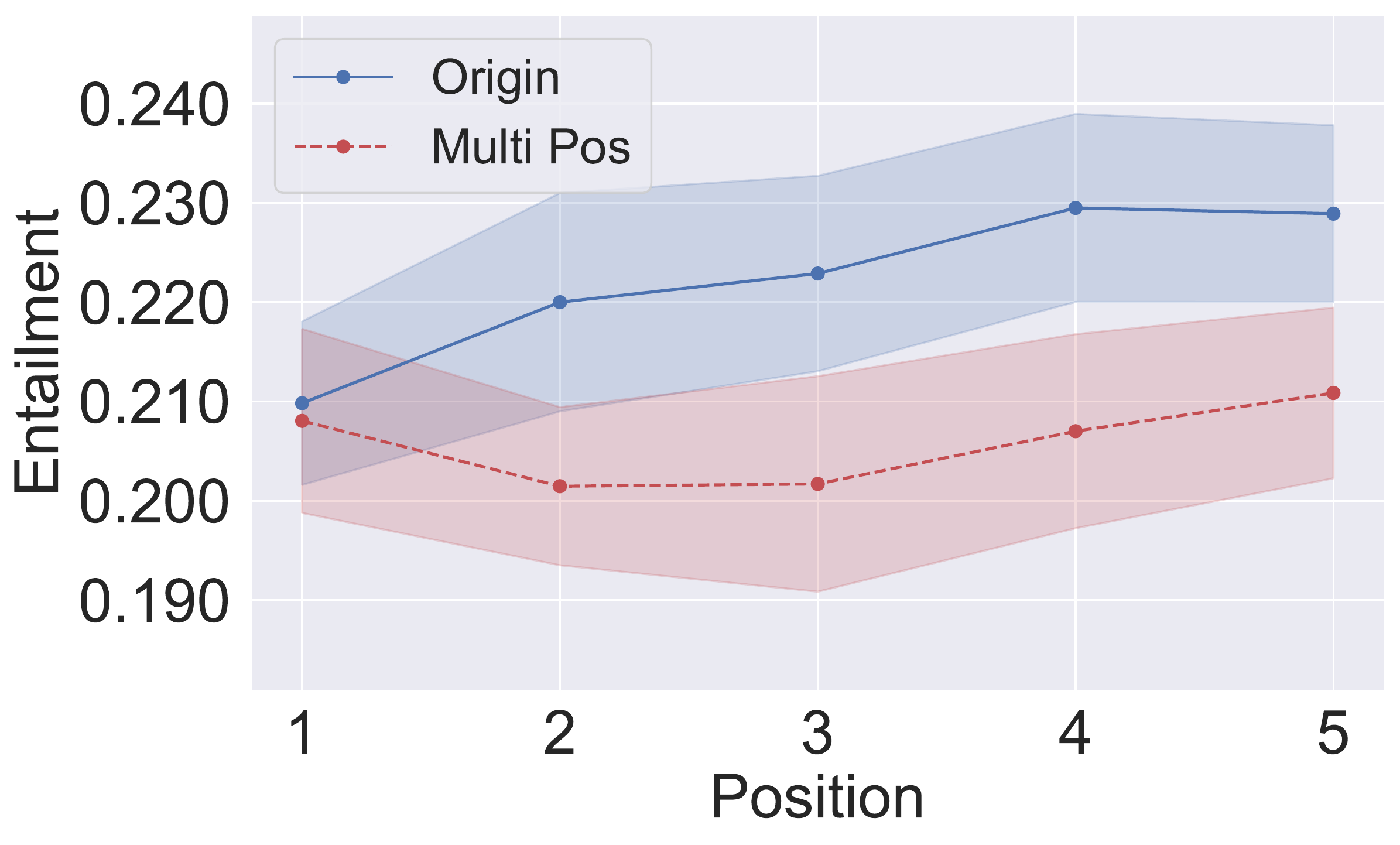}
            \caption[]%
            {{\tiny GPT on Persona-Chat}}    
            \label{fig:gpt_5_p}
        \end{subfigure}
        \hfill
        \begin{subfigure}[b]{0.245\linewidth}   
            \centering 
            \includegraphics[width=\linewidth]{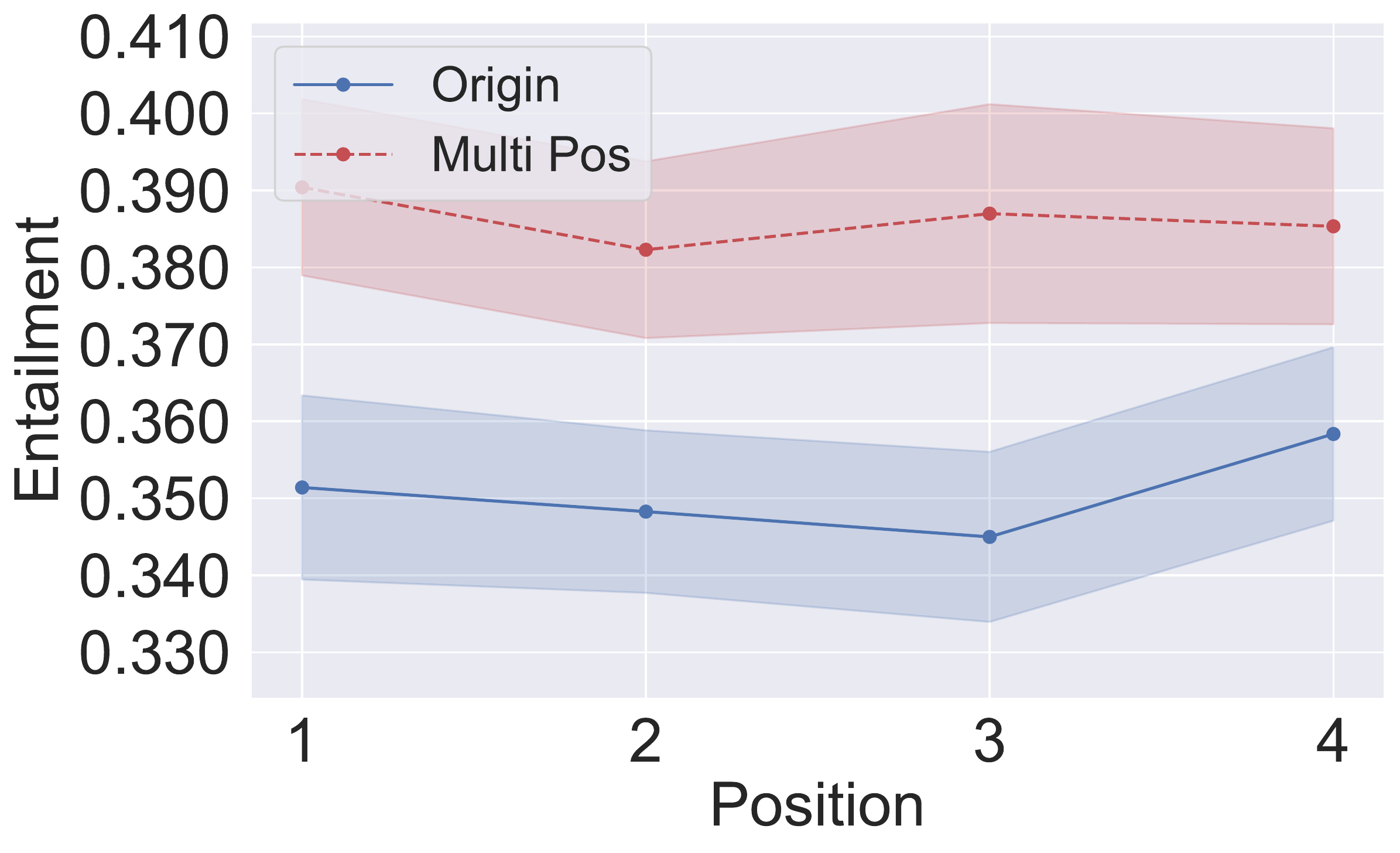}
            \caption[]%
            {{\tiny GPT on Topical-Chat}}    
            \label{fig:gpt_4_t}
        \end{subfigure}
        \hfill
        \begin{subfigure}[b]{0.245\linewidth}   
            \centering 
            \includegraphics[width=\linewidth]{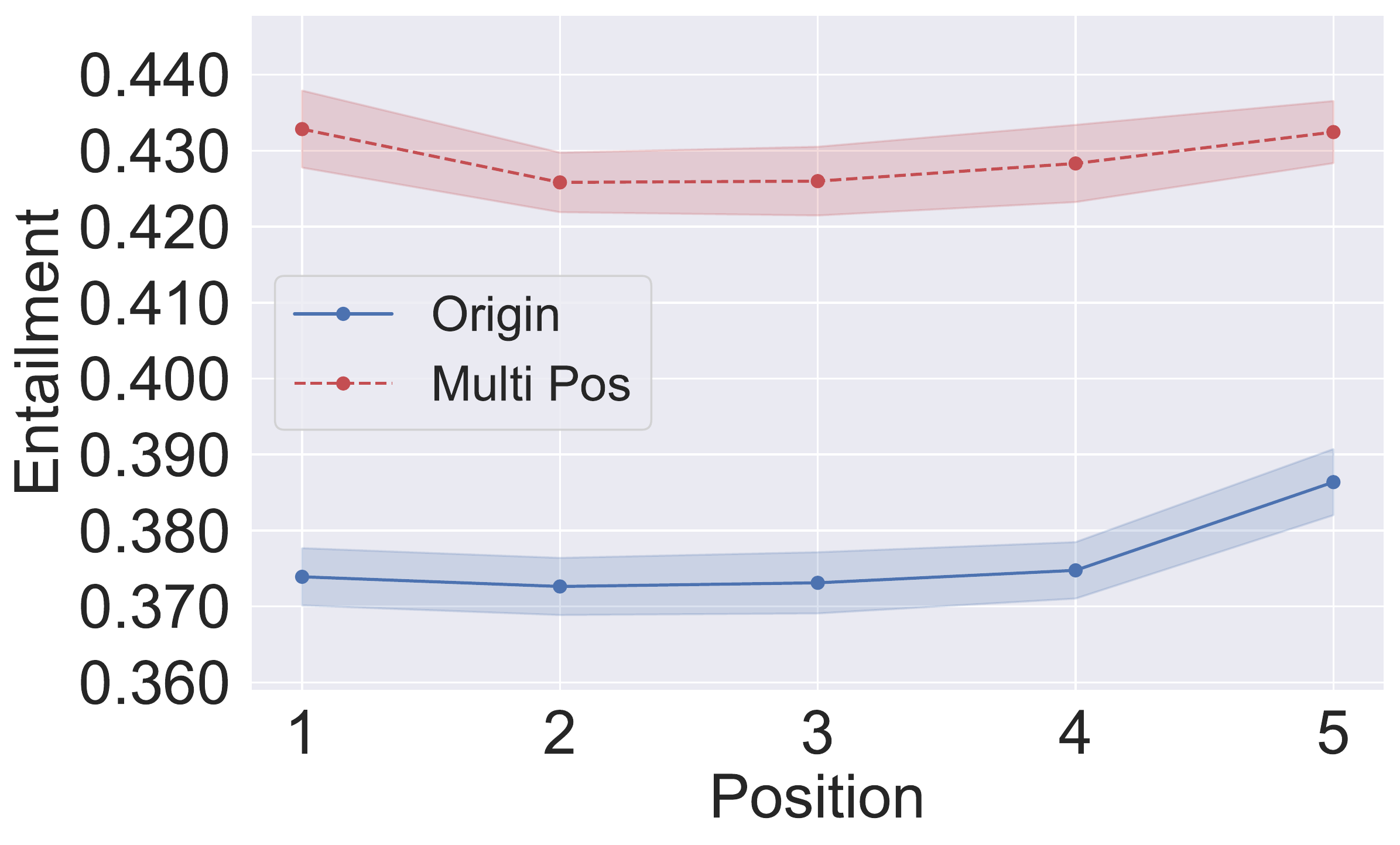}
            \caption[]%
            {{\tiny GPT on Topical-Chat}}
            \label{fig:gpt_5_t}
        \end{subfigure}
        \vskip\baselineskip
        \begin{subfigure}[b]{0.245\linewidth}
            \centering
            \includegraphics[width=\linewidth]{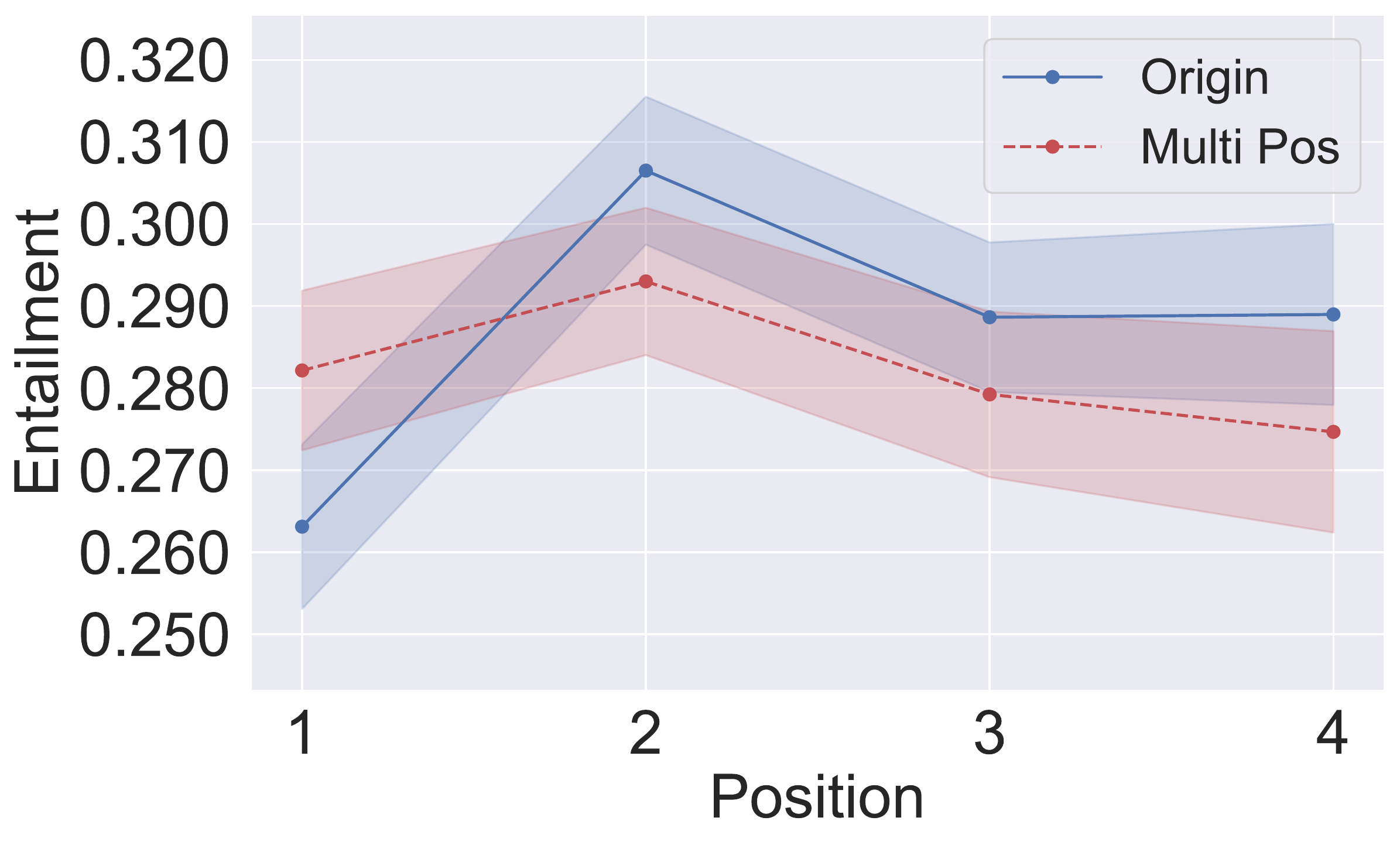}
            \caption[]%
            {{\tiny GPT-2 on Persona-Chat}}    
            \label{fig:gpt2_4_p}
        \end{subfigure}
        \hfill
        \begin{subfigure}[b]{0.245\linewidth}  
            \centering 
            \includegraphics[width=\linewidth]{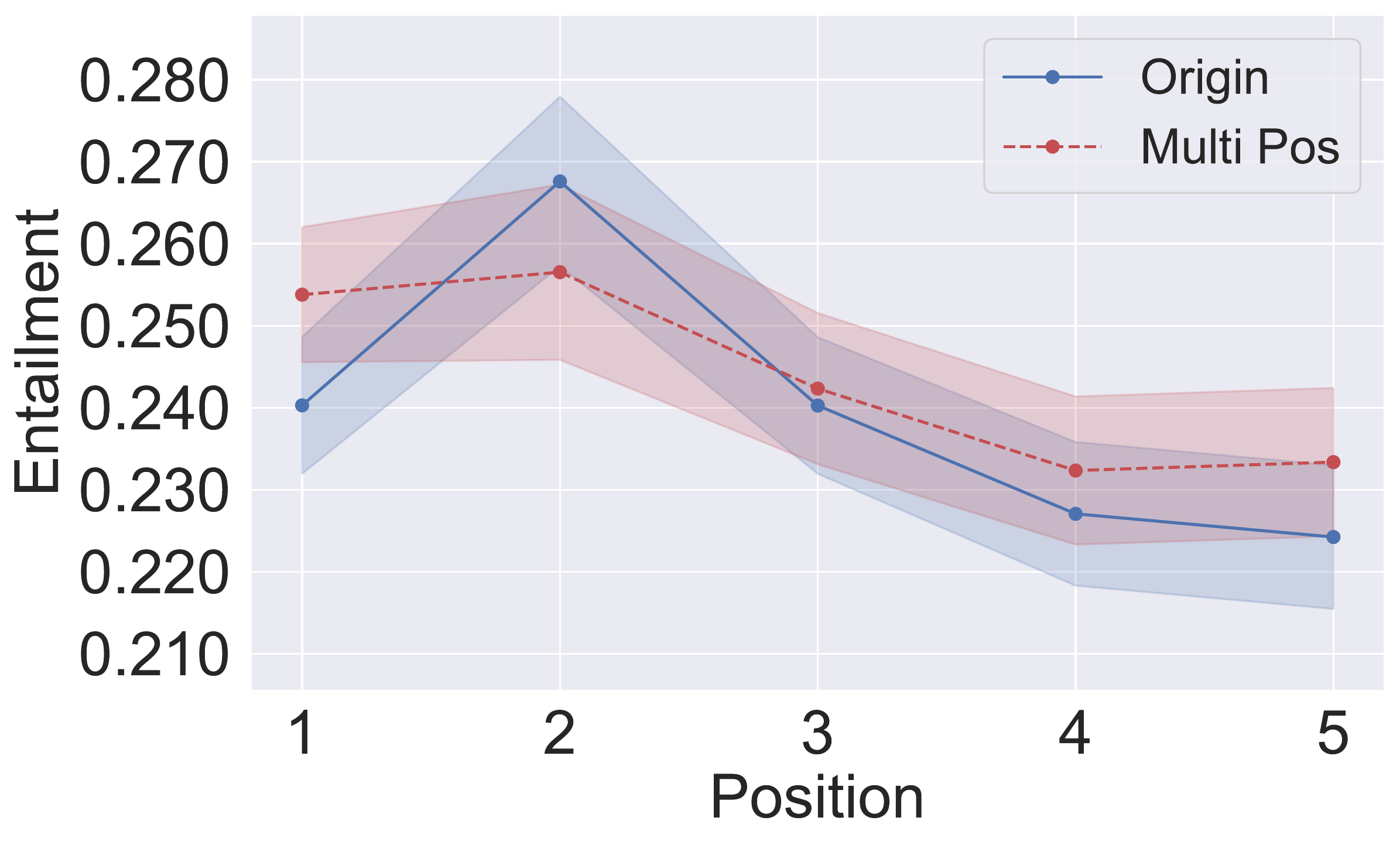}
            \caption[]%
            {{\tiny GPT-2 on Persona-Chat}}    
            \label{fig:gpt2_5_p}
        \end{subfigure}
\hfill
        \begin{subfigure}[b]{0.245\linewidth}   
            \centering 
            \includegraphics[width=\linewidth]{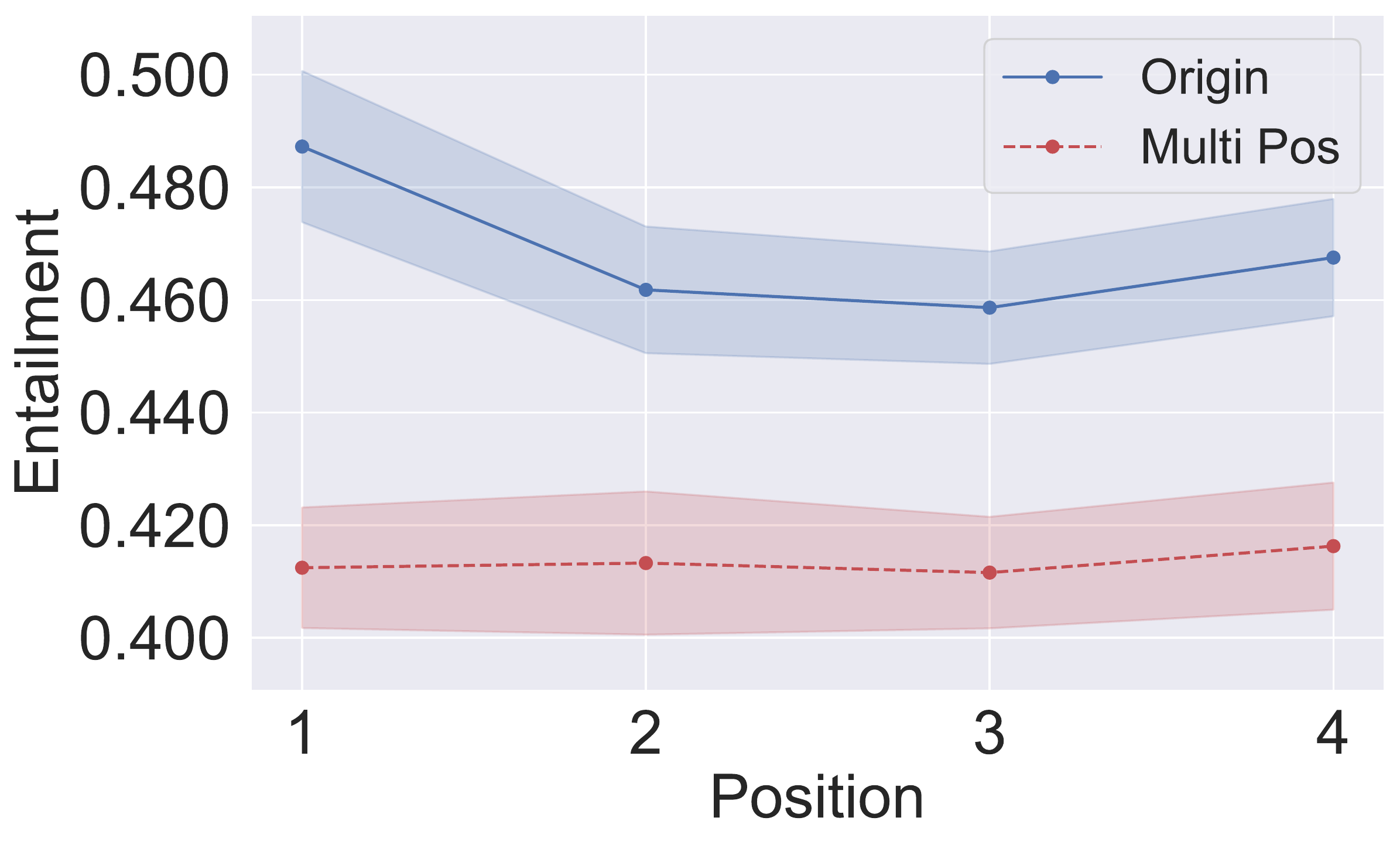}
            \caption[]%
            {{\tiny GPT-2 on Topical-Chat}}    
            \label{fig:gpt2_4_t}
        \end{subfigure}
        \hfill
        \begin{subfigure}[b]{0.245\linewidth}   
            \centering 
            \includegraphics[width=\linewidth]{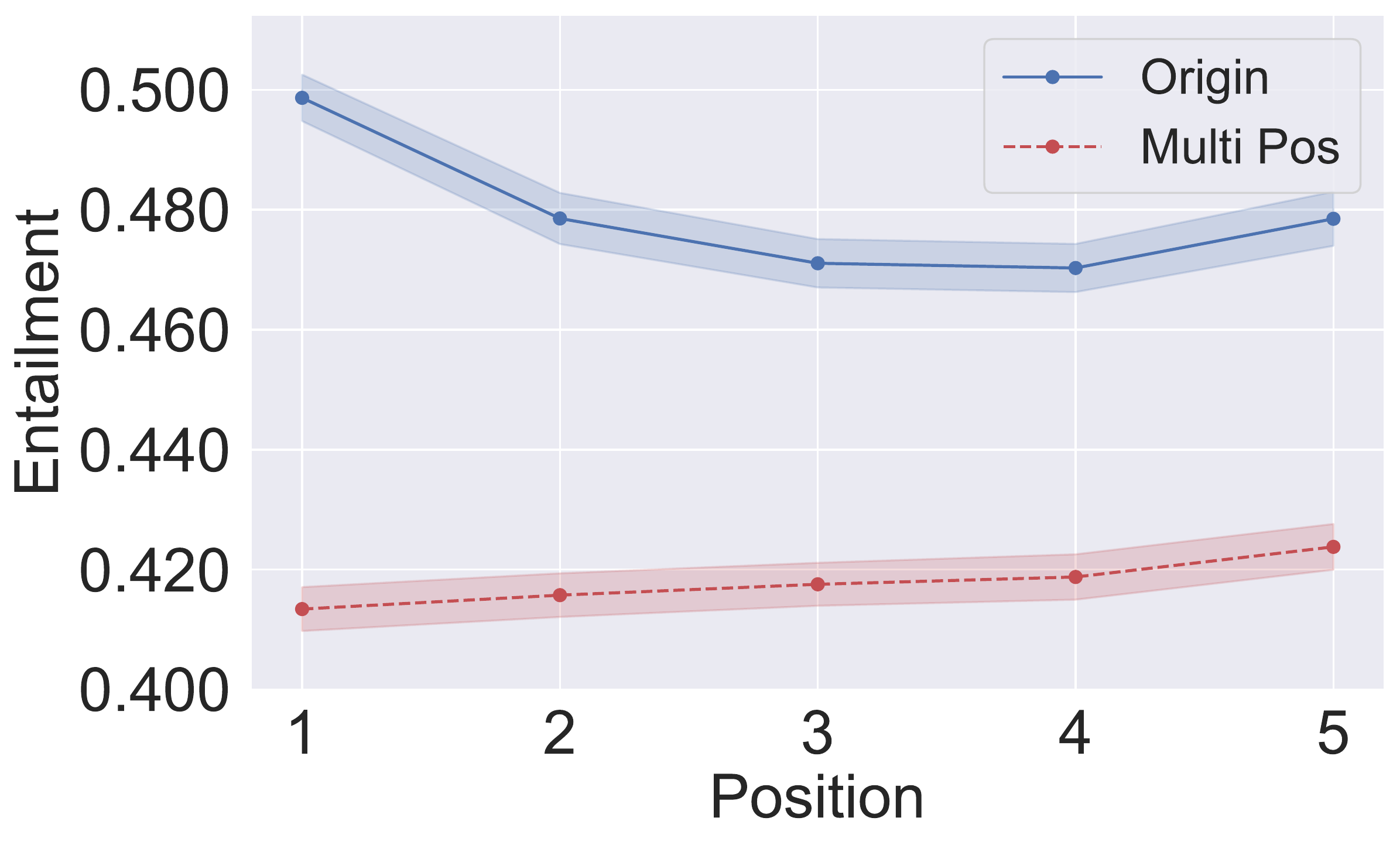}
            \caption[]%
            {{\tiny GPT-2 on Topical-Chat}}
            \label{fig:gpt2_5_t}
        \end{subfigure}
        \caption{Experimental results under TransferTransfo method, the lines indicate the average of 50 times shuffling results with standard deviation represented in the area. The data with 4 and 5 knowledge sets are displayed separately.}
        \label{fig:origin_gpt_gpt2}

        \begin{subfigure}[b]{0.245\linewidth}
            \centering
            \includegraphics[width=\linewidth]{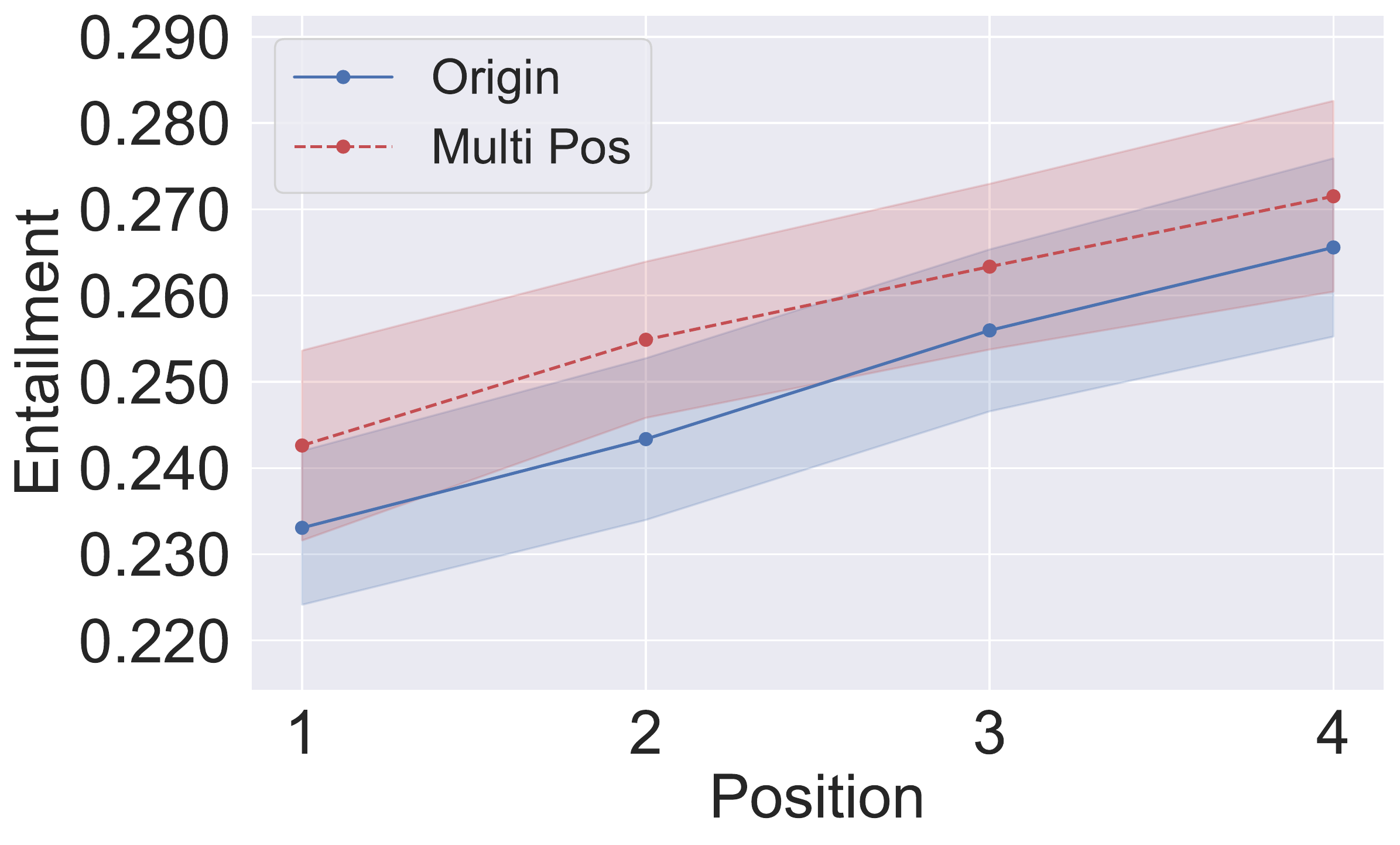}
            \caption[]%
            {{\tiny GPT on Persona-Chat}}    
            \label{fig:gpt_4_p_m}
        \end{subfigure}
        \hfill
        \begin{subfigure}[b]{0.245\linewidth}  
            \centering 
            \includegraphics[width=\linewidth]{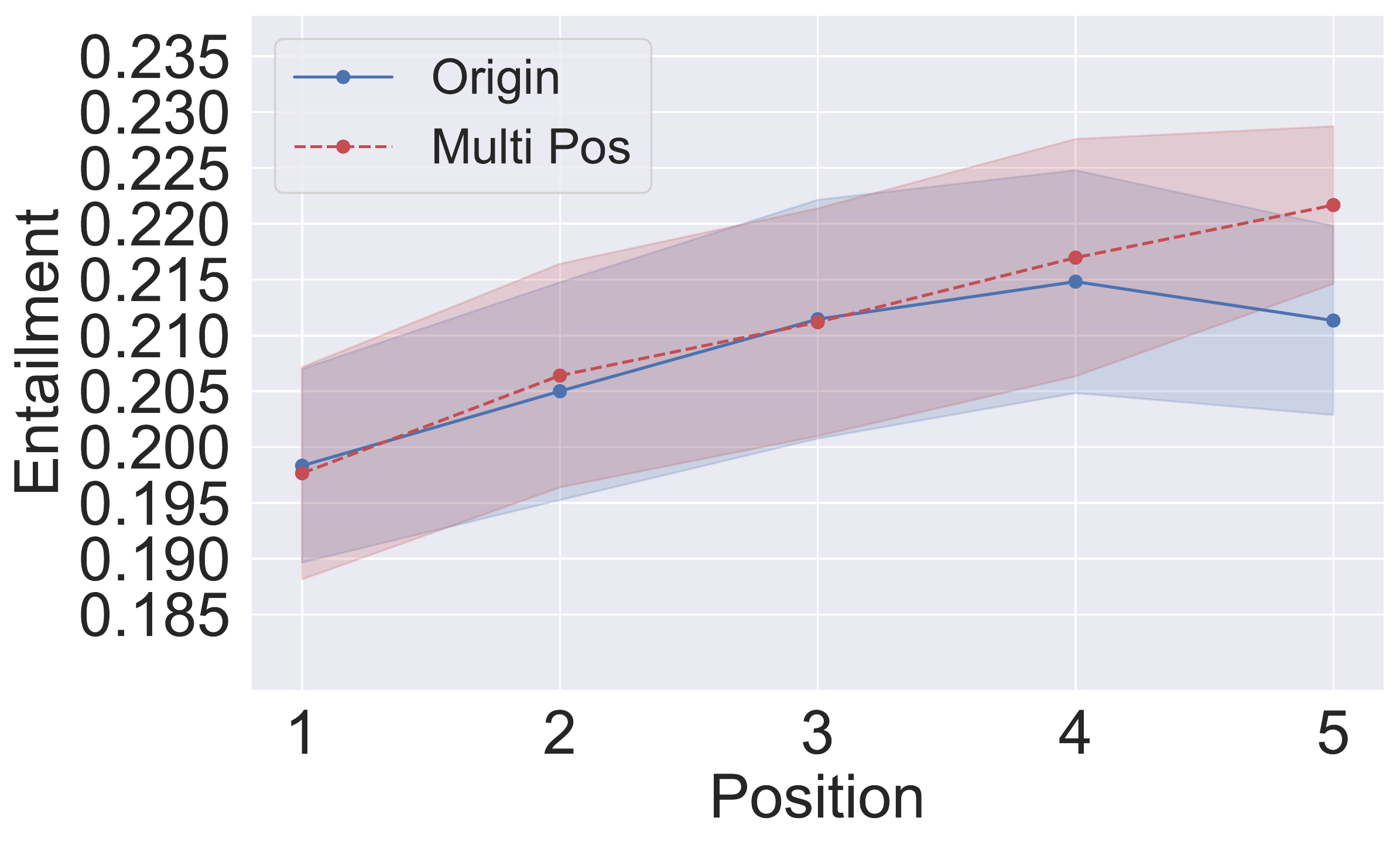}
            \caption[]%
            {{\tiny GPT on Persona-Chat}}    
            \label{fig:gpt_5_p_m}
        \end{subfigure}
        \hfill
        \begin{subfigure}[b]{0.245\linewidth}   
            \centering 
            \includegraphics[width=\linewidth]{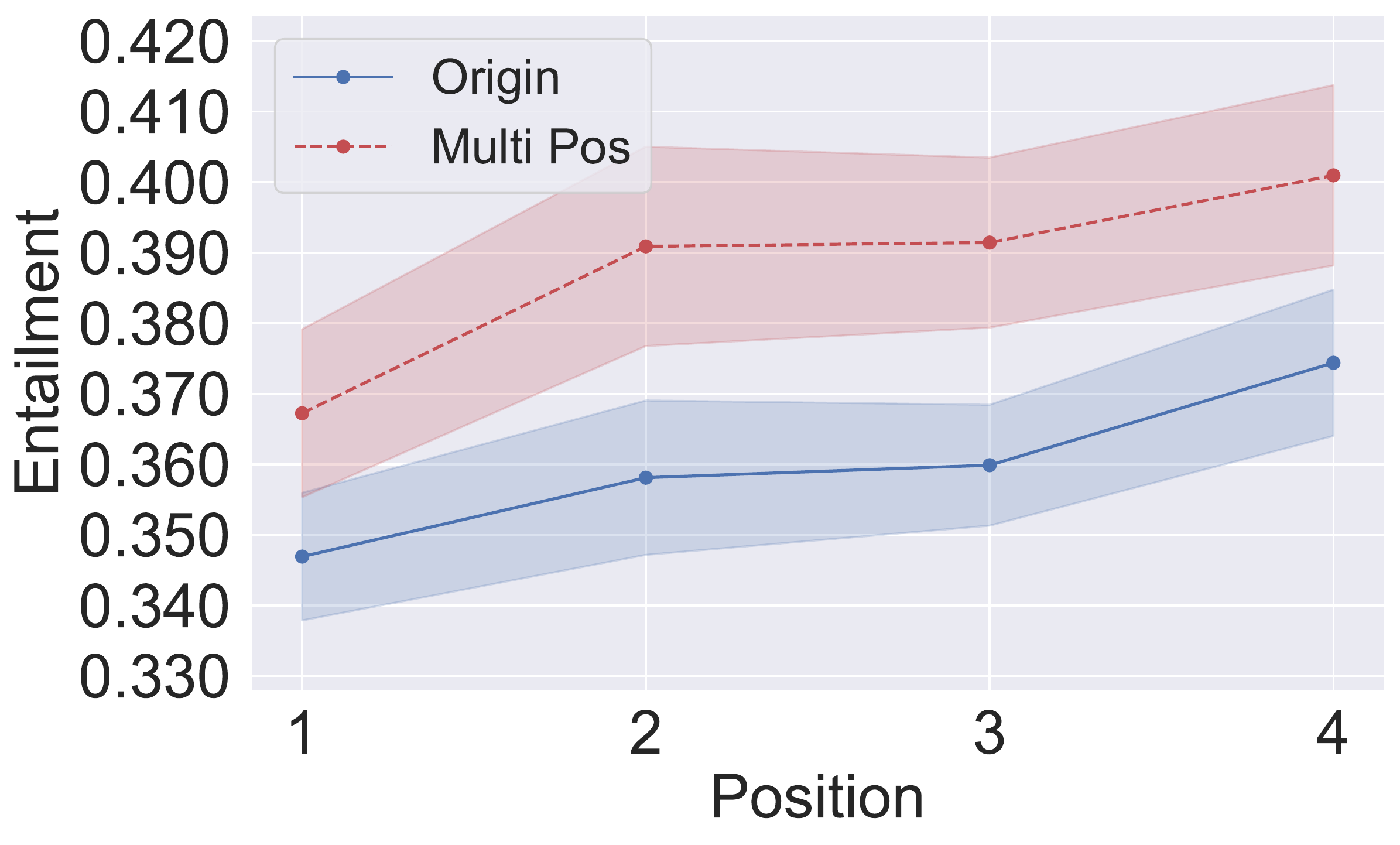}
            \caption[]%
            {{\tiny GPT on Topical-Chat}}    
            \label{fig:gpt_4_t_m}
        \end{subfigure}
        \hfill
        \begin{subfigure}[b]{0.245\linewidth}   
            \centering 
            \includegraphics[width=\linewidth]{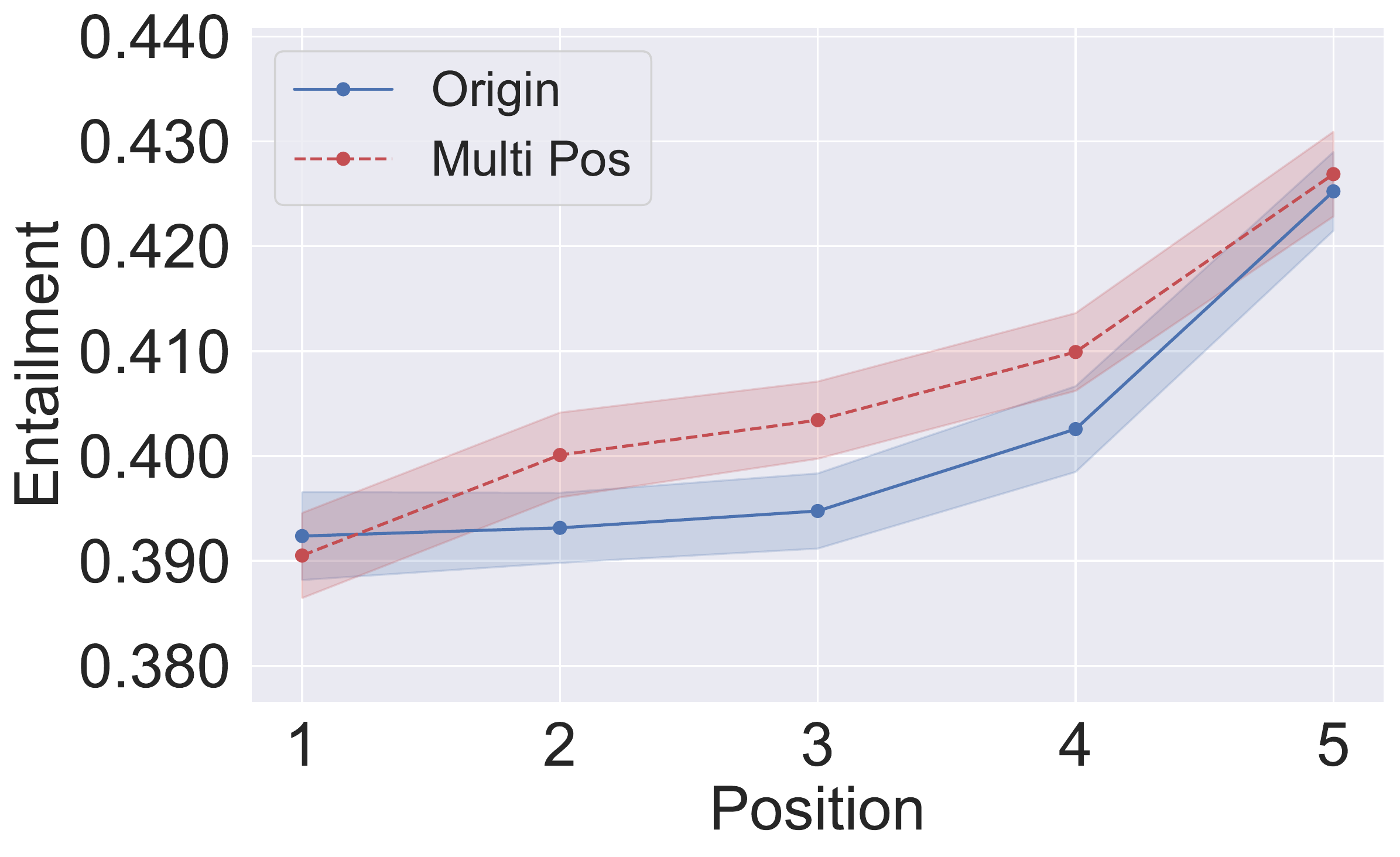}
            \caption[]%
            {{\tiny GPT on Topical-Chat}}
            \label{fig:gpt_5_t_m}
        \end{subfigure}
        \vskip\baselineskip
        \begin{subfigure}[b]{0.245\linewidth}
            \centering
            \includegraphics[width=\linewidth]{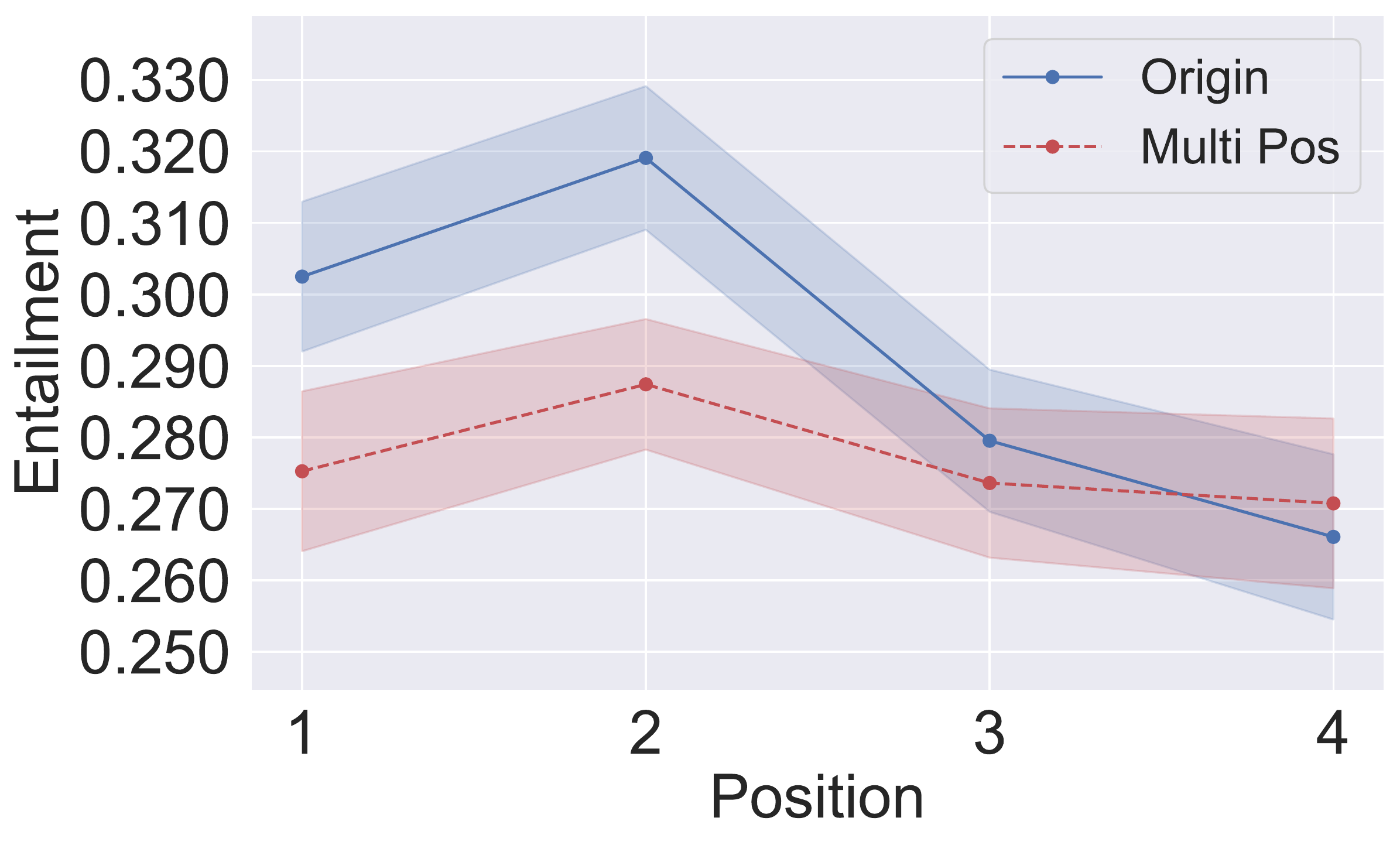}
            \caption[]%
            {{\tiny GPT-2 on Persona-Chat}}    
            \label{fig:gpt2_4_p_m}
        \end{subfigure}
        \hfill
        \begin{subfigure}[b]{0.245\linewidth}  
            \centering 
            \includegraphics[width=\linewidth]{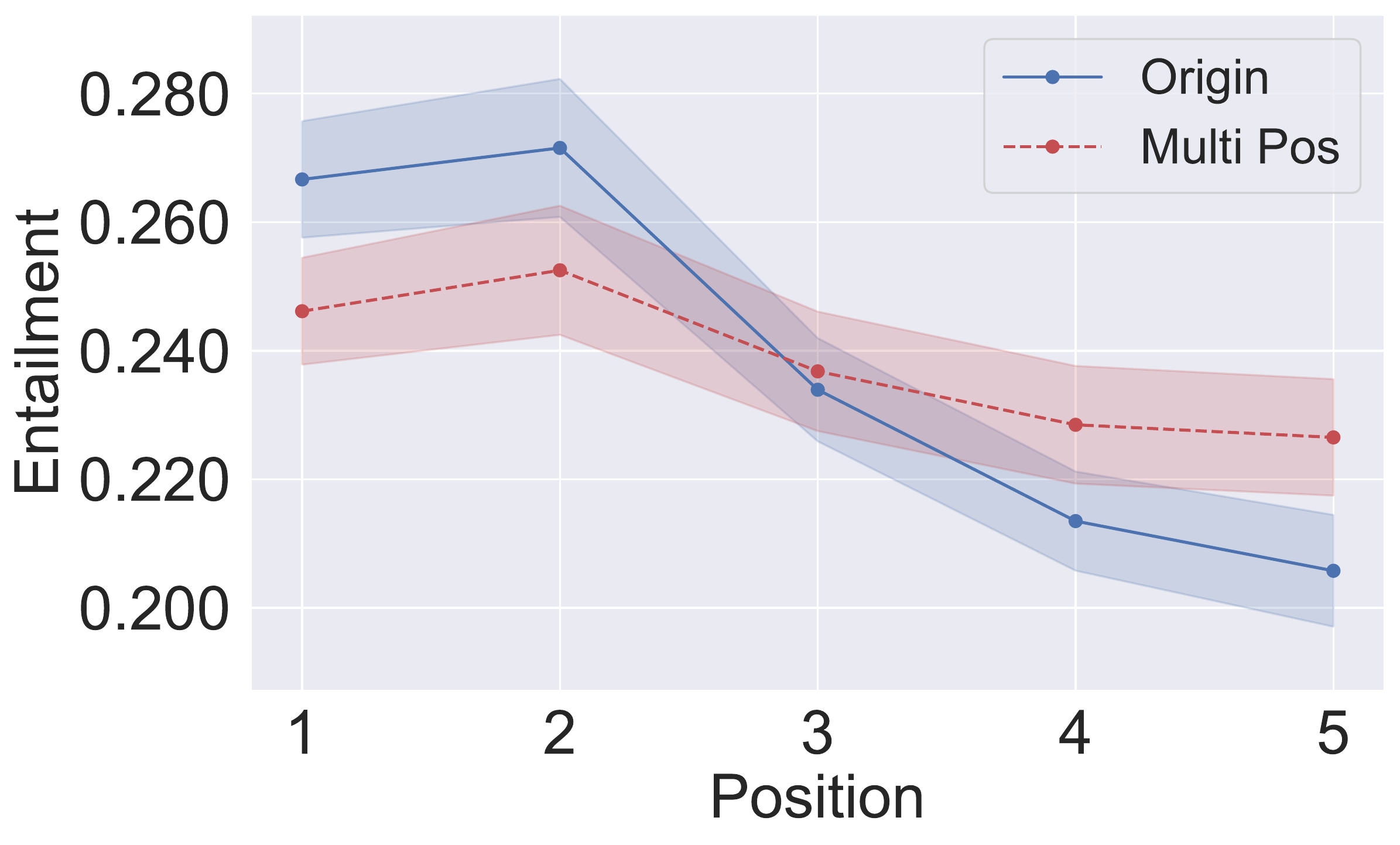}
            \caption[]%
            {{\tiny GPT-2 on Persona-Chat}}    
            \label{fig:gpt2_5_p_m}
        \end{subfigure}
        \hfill
        \begin{subfigure}[b]{0.245\linewidth}   
            \centering 
            \includegraphics[width=\linewidth]{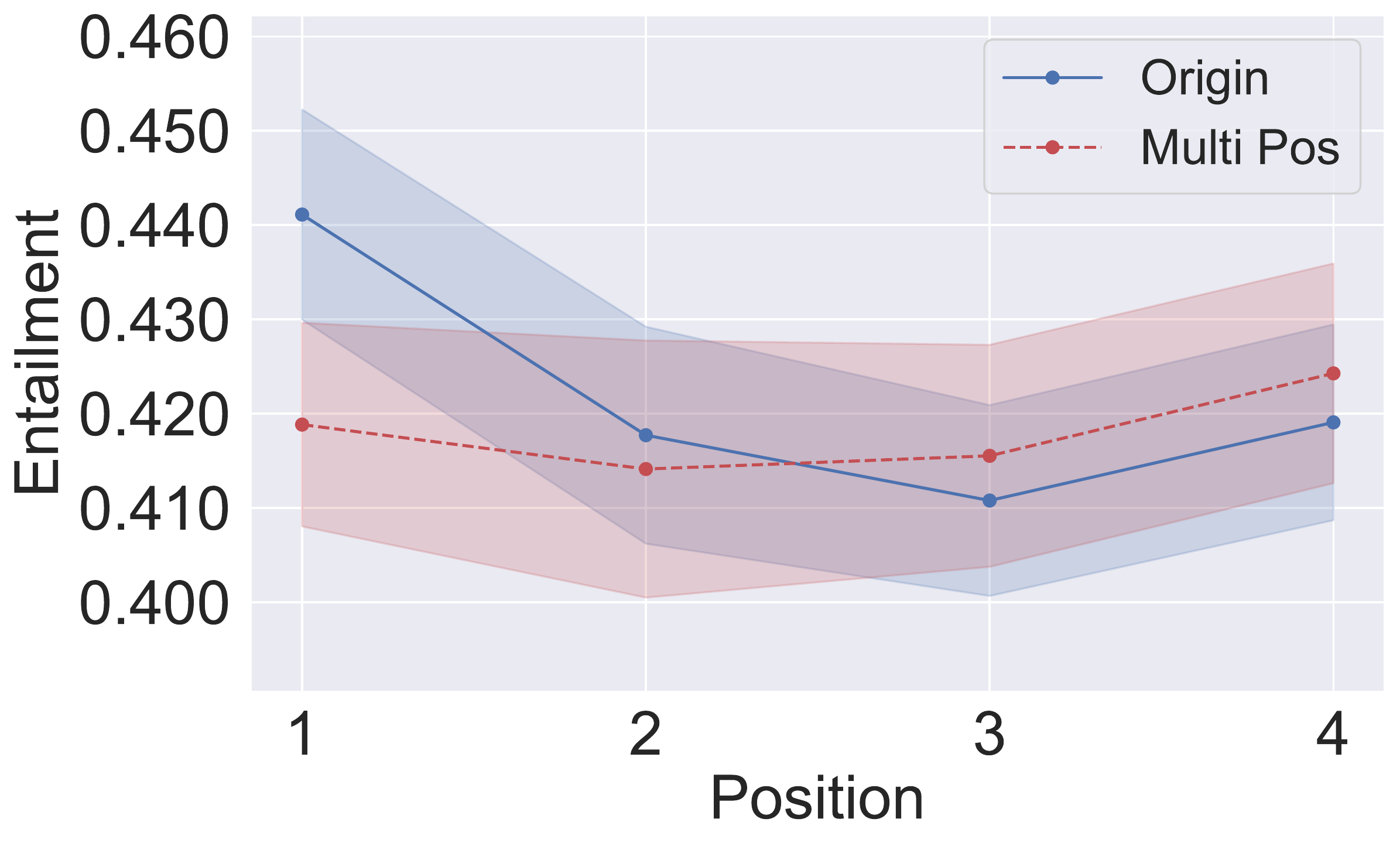}
            \caption[]%
            {{\tiny GPT-2 on Topical-Chat}}    
            \label{fig:gpt2_4_t_m}
        \end{subfigure}
        \hfill
        \begin{subfigure}[b]{0.245\linewidth}   
            \centering 
            \includegraphics[width=\linewidth]{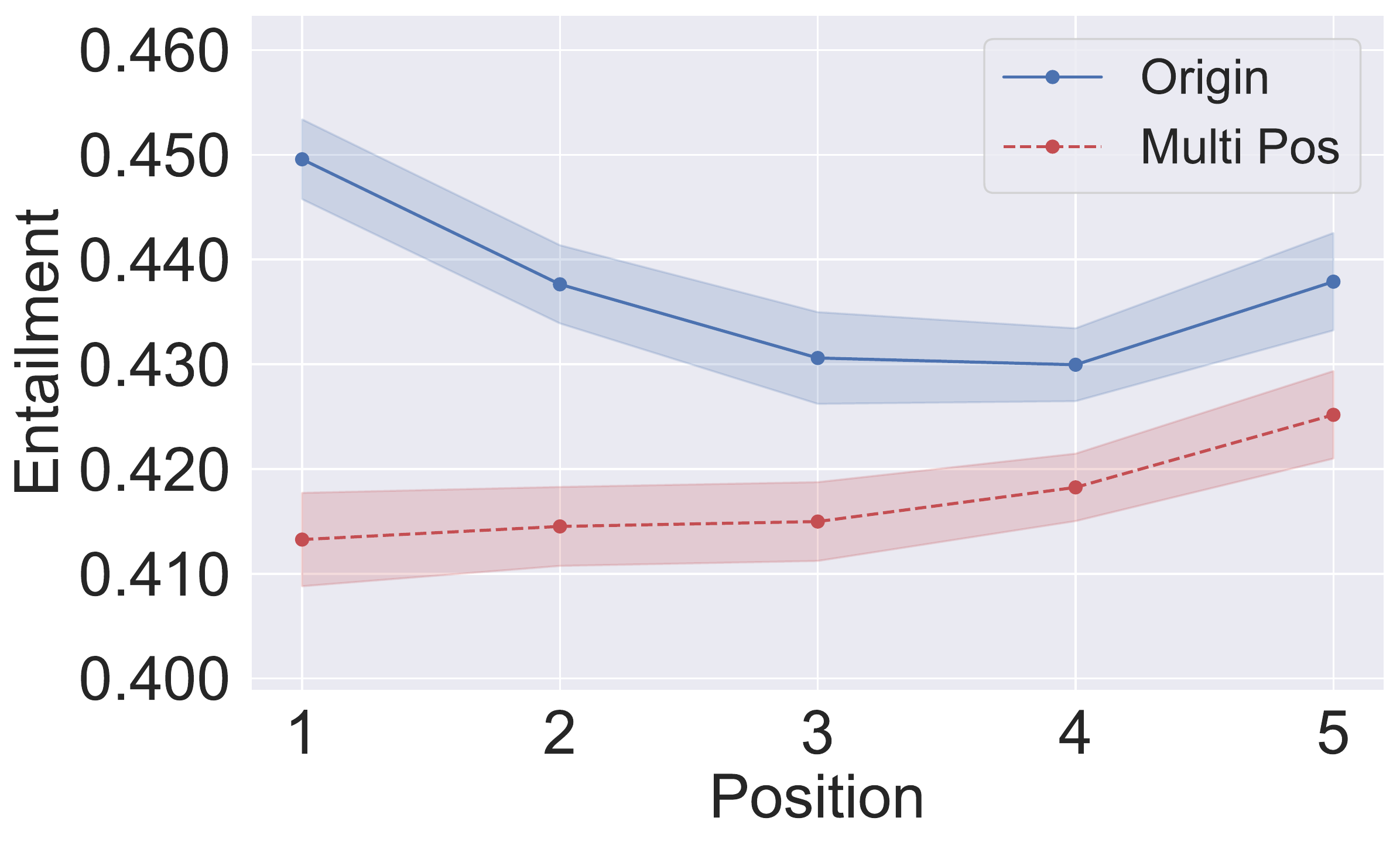}
            \caption[]%
            {{\tiny GPT-2 on Topical-Chat}}
            \label{fig:gpt2_5_t_m}
        \end{subfigure}
        \caption{Experimental results under LM loss only method, the lines indicate the average of 50 times shuffling results with standard deviation represented in the area. The data with 4 and 5 knowledge sets are displayed separately.}
        \label{fig:origin_gpt_gpt2_m}
    \end{figure*}

\subsection{Experimental Setups}
We conduct experiments on two datasets:
\\\textbf{Persona-Chat} \cite{zhang-etal-2018-personalizing}: This persona-grounded dialogue dataset consists of crowdsourced dialogues between a pair of annotators provided with 4-5 persona statements each.
\\\textbf{Topical-Chat} \cite{Gopalakrishnan2019}: This is a knowledge-grounded dialogue dataset, where the dialogs are constructed by a pair of annotators conversing about specific topics. The annotators are provided with wiki data with 4-5 facts as knowledge sources. 

In our experimental setup, we shuffle the knowledge set's order 50 times during testing and implement TransferTransfo on GPT \cite{radford2018improving} and GPT-2 \cite{radford2019language} models.

\section{The Order Effect of the Knowledge Set}
Models are said to have an order effect of input if the generated responses are sensitive and influenced by order of input sequence. Previous works \cite{sankar-etal-2019-neural, oconnor-andreas-2021-context, sinha-etal-2021-masked, Lampinen2022CanLM, Webson2021DoPM, Xu2020ATO, Khandelwal2018SharpNF} focus on whether perturbation in dialogue history affect models' responses. 
In this work, to be more specific, we investigate if sentence level change in the order of input knowledge sets will result in substantial semantic differences in the generated responses. 

\subsection{The Order Effect Measurement}
To address the sentence-level order effect of the input knowledge set in models, we aim to measure the semantic difference given different orders of knowledge sentences. It is intuitive to measure if the response content is influenced by knowledge sets order. In other words, we measure the distribution of response-knowledge relationship in different positions.
We build a Natural Language Inference (NLI) classifier to evaluate the degree of entailment between responses and each knowledge in the set.

\begin{table}[htp!]
\centering

\begin{adjustbox}{max width=\columnwidth}
\Huge
\begin{tabular}{cccccc}
\hline

\multicolumn{1}{c|}{Model} & \multicolumn{1}{c|}{Method} & \multicolumn{2}{c|}{Persona} & \multicolumn{2}{c}{Topical} \\ \hline
 &  & TT. & \multicolumn{1}{c|}{LM.} & TT. & LM. \\ \hline
 &  & \multicolumn{4}{c}{Entailment  Max - Min} \\ \hline
\multicolumn{1}{c|}{\multirow{2}{*}{GPT}} & \multicolumn{1}{c|}{Origin} & .048 / .037 & \multicolumn{1}{c|}{.052 / .035} & .037 / .022 & .046 / .041 \\
\multicolumn{1}{c|}{} & \multicolumn{1}{c|}{Multi Pos} & \textbf{.023 / .028} & \multicolumn{1}{c|}{.051 / .041} & \textbf{.031 / .016} & .058 / .044 \\
\multicolumn{1}{c|}{\multirow{2}{*}{GPT-2}} & \multicolumn{1}{c|}{Origin} & .062 / .062 & \multicolumn{1}{c|}{.075 / .085} & .052 / .036 & .052 / .027 \\
\multicolumn{1}{c|}{} & \multicolumn{1}{c|}{Multi Pos} & \textbf{.039 / .044} & \multicolumn{1}{c|}{\textbf{.038 / .045}} & \textbf{.027 / .018} & \textbf{.035 / .021} \\ \hline
 &  & \multicolumn{4}{c}{Perplexity $\downarrow$} \\ \hline
\multicolumn{1}{c|}{\multirow{2}{*}{GPT}} & \multicolumn{1}{c|}{Origin} & 52.29 & \multicolumn{1}{c|}{54.31 } & 39.31 & 36.80   \\
\multicolumn{1}{c|}{} & \multicolumn{1}{c|}{Multi Pos} & 55.47  & \multicolumn{1}{c|}{58.43  } & 42.37 & 42.98 \\
\multicolumn{1}{c|}{\multirow{2}{*}{GPT-2}} & \multicolumn{1}{c|}{Origin} & 61.69  & \multicolumn{1}{c|}{61.80 } & 20.50 & 18.84  \\
\multicolumn{1}{c|}{} & \multicolumn{1}{c|}{Multi Pos} & 60.18 & \multicolumn{1}{c|}{58.91 } & 17.40  & 17.30 \\ \hline
 &  & \multicolumn{4}{c}{Coherence} \\ \hline
\multicolumn{1}{c|}{\multirow{2}{*}{GPT}} & \multicolumn{1}{c|}{Origin} & 0.633 & \multicolumn{1}{c|}{0.636 } & 0.793 & 0.770 \\
\multicolumn{1}{c|}{} & \multicolumn{1}{c|}{Multi Pos} & 0.644 & \multicolumn{1}{c|}{0.621 } & 0.732 & 0.744 \\
\multicolumn{1}{c|}{\multirow{2}{*}{GPT-2}} & \multicolumn{1}{c|}{Origin} & 0.661 & \multicolumn{1}{c|}{0.667 } & 0.840 & 0.843 \\
\multicolumn{1}{c|}{} & \multicolumn{1}{c|}{Multi Pos} & 0.648 & \multicolumn{1}{c|}{0.662 } & 0.830 & 0.831 \\ \hline
 &  & \multicolumn{4}{c}{Diverstiy $\downarrow$} \\ \hline
\multicolumn{1}{c|}{\multirow{2}{*}{GPT}} & \multicolumn{1}{c|}{Origin} & 0.815 & \multicolumn{1}{c|}{0.822 } & 0.844 & 0.846 \\
\multicolumn{1}{c|}{} & \multicolumn{1}{c|}{Multi Pos} & 0.821 & \multicolumn{1}{c|}{0.833 } & 0.870 & 0.862 \\
\multicolumn{1}{c|}{\multirow{2}{*}{GPT-2}} & \multicolumn{1}{c|}{Origin} & 0.808 & \multicolumn{1}{c|}{0.811 } & 0.833 & 0.833 \\
\multicolumn{1}{c|}{} & \multicolumn{1}{c|}{Multi Pos} & 0.816 & \multicolumn{1}{c|}{0.817 } & 0.843 & 0.845 \\ \hline
\end{tabular}
\end{adjustbox}
    \caption{\label{tab:table1}The results of measurements. The Max-Min of entailment are reported in 4 knowledge / 5 knowledge.
    The mean of quality across 50 runs are reported and standard deviation are reported in Appendix \ref{std}.}
\end{table}

The Natural Language Inference Classifier is built with BERT model \cite{devlin-etal-2019-bert}, trained on the Dialogue NLI dataset \cite{welleck-etal-2019-dialogue}, which is built on top of Persona-Chat dataset \cite{zhang-etal-2018-personalizing}. The annotators label the relationship between persona and response in Persona-Chat with entail, neutral, and contradict classes. 

\subsection{Results and Discussions for Order Effect}
Figures \ref{fig:origin_gpt_gpt2} and \ref{fig:origin_gpt_gpt2_m} show the entailment scores of the response with each position of knowledge. Figure \ref{fig:origin_gpt_gpt2} presents the experiments of TransferTransfo with GPT and GPT-2 models across Persona-Chat and Topical-Chat datasets.
Figure \ref{fig:origin_gpt_gpt2_m} shows the results with "LM Loss only Method", which refers to TransferTransfo without the next sentence prediction.
We observe that the distribution of data containing only four knowledge statements is very different compared to data containing five knowledge statements. Hence we show them separately. 

The NLI classification results are shown with BLUE lines. We can see that the distribution of entailment scores on different positions are imbalanced.
In the experiments on the GPT model, (figures \ref{fig:gpt_4_p}, \ref{fig:gpt_5_p}, \ref{fig:gpt_4_t}, \ref{fig:gpt_5_t}, \ref{fig:gpt_4_p_m}, \ref{fig:gpt_5_p_m}, \ref{fig:gpt_4_t_m}, and \ref{fig:gpt_5_t_m}), it can be observed under both TransferTransfo and LM loss only methods, the entailment score on the last position is always the highest. In fact, there is a huge gap between the entailment scores with the first knowledge and the last knowledge statements. This indicates that GPT model focuses more on 
the last position of knowledge.

However, the behavior of GPT-2 is very different from GPT model. From Figures \ref{fig:gpt2_4_p}, \ref{fig:gpt2_5_p}, \ref{fig:gpt2_4_t}, \ref{fig:gpt2_5_t}, \ref{fig:gpt2_4_p_m}, \ref{fig:gpt2_5_p_m}, \ref{fig:gpt2_4_t_m}, and \ref{fig:gpt2_5_t_m}, we can see that GPT-2 models focus more on the earlier knowledge statements in the sequence rather than the later ones.

These results show that the order effect exists across GPT and GPT-2 models (although different) and is influencing models' responses and this needs to be solved.

\section{Alleviate the Order Effect}
In this section, we analyse the reason for the order effect in the GPT-series models
and propose a method to alleviate the phenomenon. Figure \ref{fig:framework} shows the input format of the classic GPT-series. There are three types of embeddings in the model: word embedding to capture the semantic meaning of each word, token embedding to represent the speaker 
and absolute position embedding that encodes position information of input sequence.

Figure \ref{fig:framework} shows that the position ids for each knowledge start from zero with different positional embedding layers. 
In this naive setting, knowledge of the set are treated equally and not input with the order  during training.



\subsection{Results and Discussion}

In the same Figures \ref{fig:origin_gpt_gpt2} and  \ref{fig:origin_gpt_gpt2_m}, the RED lines demonstrate the entailment result after applying multiple position embedding. We observe that all the red lines, which are the GPT-series applied multiple position embeddings, are much smoother compared to BLUE lines in both figures. Furthermore, we report the difference between maximum and  minimum entailment across the positions in Table \ref{tab:table1}. It shows that the difference is negligible after applying multiple position embeddings.
This indicates that we can alleviate the order effect under models trained with with multiple position embedding. However, we also observed that on Figure \ref{fig:origin_gpt_gpt2_m} some red lines are still as steep as before, which means the order effect still exists. 
We think that the model trained only with LM loss treats knowledge like history and does not ground models on knowledge sets. Under this scenario, the multiple position embedding doesn't work well.

For the measurement of quality, Table \ref{tab:table1} shows the perplexity, coherence, and diversity. The details are included in Appendix \ref{metrics}. We found tiny drops between origin and multiple position embedding. More specifically, our proposed method does not crash the models and can still make models generate plausible responses.

\section{Conclusions}
In this paper, we investigate whether the order of knowledge set will influence dialogue models' responses. Our experiments across several datasets show that the GPT-series models unfairly pay attention to the knowledge set and are influenced by order of knowledge. To solve this problem, we study the reason for the phenomenon and propose simple method to alleviate the order effect in models. The experimental results show that our approach reduces the order effect and makes the model select the knowledge uniformly.

\section*{Limitations}
This work has potential limitations:
\begin{itemize}

    \item We found that on the Figure \ref{fig:origin_gpt_gpt2} and \ref{fig:origin_gpt_gpt2_m}, The entailment of the methods after applying multiple position embedding (RED lines) are sometimes lower than  origin methods(BLUE lines). This is not meet our expectations since we don't want our method to decrease performance. In our opinion, we think the reason might be the embedding method has never been seen before during the pretraining of models, which requires the model’s additional efforts to adapt the embedding, thus hurts the performance.. We leave it as future work to be improved.
    \item We also found that the multiple position embedding does not work very well to alleviate the order effect in the LM loss-only settings\ref{fig:origin_gpt_gpt2_m}. We have discussed this in previous sections. Since LM loss only does not help the model distinguish which parts in the input sequence are knowledge set and thus treat them the same as history. The multiple position embedding will not be trained finely to help the model distinguish. We also left this as a future work to be improved.
\end{itemize}


\bibliography{anthology,custom}
\bibliographystyle{acl_natbib}

\appendix

\section{Appendix}
\label{sec:appendix}
\subsection{Experimental Details}
\begin{itemize}
    \item \textbf{Hyperparameters}: For the  Hyperparameters we use to conduct experiments, we follow TransferTransfo link \url{https://github.com/huggingface/transfer-learning-conv-ai}. They obtain these Hyperparameters by grid searching.
    More specifically, They finetuned the model with a batch
size of 32 sequences , and finetune the models approximately 2 epochs over training dataset. They used Adam with a learning rate of 6.25e-5,
and a coefficient of 2 on the LM loss when summing with the next-sentence prediction loss . The learning rate was linearly decayed to zero over the course of the training.
\item \textbf{Datasets}: The link to download Persona-Chat \url{https://parl.ai/docs/tasks.html#persona-chat} and the train/valid/test split is 9907/1000/968 dialogues..
For the link to download Topical-Chat \url{https://github.com/alexa/Topical-Chat} and the train/valid/test split is 8628/1078/1078 dialogues.
\item \textbf{Pretrained Models}: For GPT model we use gpt-medium as our pretrain model and use microsoft/DialoGPT-medium as initial checkpoint for GPT-2 model.
    
\end{itemize}
\subsection{Evaluation Metrics} \label{metrics}
In addition to entailment, we aimed to employ other metrics that are also important to measure a dialogue system.
\\\textbf{Perplexity} \citep{chen1998evaluation}: Here we employed the pretrained GPT-2 language model $GPT$ to judge if the output sentence $C(x)$ was an acceptable sentence. The computation of Perplexity \citep{chen1998evaluation} is shown below.
\begin{equation}
  PPL = \prod_{i=1}^{T} \frac{1}{(GPT(C(x, D)_i|x))^{1/T}}
\end{equation}
\\\textbf{Coherence}: We employed the DialogRPT \citep{gao2020dialogrpt} to calculate the coherence between conversation model's output and the input context. DialogRPT \citep{gao2020dialogrpt} is a GPT2-based ranker that finetuned on 133M human feedback data. With the contrastive learning approach that DialogRPT used. The ranker has better understanding on how relevant the response is for the given context. In our evaluation, we take the the probability that output by DialogRPT coherence model (\textit{human\_vs\_rand}) as our coherence metric.
\\\textbf{Diversity}: BLEU score \cite{papineni2002bleu} is a commonly used metric for automatically evaluating machine translation. However, the Self-BLEU \cite{zhu2018texygen} score here was applied to measure the diversity of chatbot responses. Regarding one sentence as the prediction and the others as the reference, we can calculate BLEU score for every sentence, and the average is the Self-BLEU score. A lower Self-BLEU score implies more diversity of the chatbot responses.

\subsection{Standard Deviation of Quality Metrics} \label{std}
\begin{table}[ht!]
\centering
\begin{adjustbox}{max width=\columnwidth}
\Huge
\begin{tabular}{cccccc}
\hline
\multicolumn{1}{c|}{Model} & \multicolumn{1}{c|}{Method} & \multicolumn{2}{c|}{Persona} & \multicolumn{2}{c}{Topical} \\ \hline
 &  & TT. & \multicolumn{1}{c|}{LM.} & TT. & LM.  \\ \hline
 &  & \multicolumn{4}{c}{Perplexity} \\ \hline
\multicolumn{1}{c|}{\multirow{2}{*}{GPT}} & \multicolumn{1}{c|}{Origin} & 0.23 & \multicolumn{1}{c|}{ 0.27} & 0.20 &  0.25 \\
\multicolumn{1}{c|}{} & \multicolumn{1}{c|}{Multi Pos} &  0.22 & \multicolumn{1}{c|}{  0.26} & 0.27 & 0.22 \\
\multicolumn{1}{c|}{\multirow{2}{*}{GPT-2}} & \multicolumn{1}{c|}{Origin} &  0.31 & \multicolumn{1}{c|}{ 0.29} & 0.120 &  0.09 \\
\multicolumn{1}{c|}{} & \multicolumn{1}{c|}{Multi Pos} & 0.28 & \multicolumn{1}{c|}{ 0.23} &  0.10 &  0.110 \\ \hline
 &  & \multicolumn{4}{c}{Coherence} \\ \hline
\multicolumn{1}{c|}{\multirow{2}{*}{GPT}} & \multicolumn{1}{c|}{Origin} & 0.001 & \multicolumn{1}{c|}{ 0.001} & 0.002 & 0.002 \\
\multicolumn{1}{c|}{} & \multicolumn{1}{c|}{Multi Pos} & 0.001 & \multicolumn{1}{c|}{ 0.001} & 0.002 & 0.002 \\
\multicolumn{1}{c|}{\multirow{2}{*}{GPT-2}} & \multicolumn{1}{c|}{Origin} & 0.002 & \multicolumn{1}{c|}{ 0.001} & 0.001 & 0.001 \\
\multicolumn{1}{c|}{} & \multicolumn{1}{c|}{Multi Pos} & 0.001 & \multicolumn{1}{c|}{ 0.001} & 0.001 & 0.001 \\ \hline
 &  & \multicolumn{4}{c}{Diverstiy} \\ \hline
\multicolumn{1}{c|}{\multirow{2}{*}{GPT}} & \multicolumn{1}{c|}{Origin} &  0.002 & \multicolumn{1}{c|}{  0.002} &  0.002 &  0.002 \\
\multicolumn{1}{c|}{} & \multicolumn{1}{c|}{Multi Pos} &  0.002 & \multicolumn{1}{c|}{  0.002} &  0.002 &  0.002 \\
\multicolumn{1}{c|}{\multirow{2}{*}{GPT-2}} & \multicolumn{1}{c|}{Origin} &  0.002 & \multicolumn{1}{c|}{  0.002} &  0.002 &  0.002 \\
\multicolumn{1}{c|}{} & \multicolumn{1}{c|}{Multi Pos} &  0.002 & \multicolumn{1}{c|}{  0.002} &  0.002 &  0.001 \\ \hline
\end{tabular}
\end{adjustbox}
    \caption{\label{tab:table3}The results of quality measurements. The standard deviation across 50 runs are reported.}
\end{table}
\end{document}